\documentclass[sigconf]{acmart}
\usepackage{amsmath,amsfonts}
\usepackage{algorithmic}
\usepackage{graphicx}
\usepackage{subcaption}
\usepackage[linesnumbered,ruled,vlined]{algorithm2e}
\usepackage{color}
\usepackage{tikz}
\usepackage{enumitem}
\usepackage{framed}

\AtBeginDocument{%
  }

\setcopyright{none}

\copyrightyear{2026}
\acmYear{2026}
\setcopyright{cc}
\setcctype{by}
\acmConference[ICSE-SEIP '26]{2026 IEEE/ACM 48th International Conference on Software Engineering}{April 12--18, 2026}{Rio de Janeiro, Brazil}
\acmBooktitle{2026 IEEE/ACM 48th International Conference on Software Engineering (ICSE-SEIP '26), April 12--18, 2026, Rio de Janeiro, Brazil}
\acmPrice{}
\acmDOI{10.1145/3786583.3786871}
\acmISBN{979-8-4007-2426-8/2026/04}

\begin{document}

%%
%% The "title" command has an optional parameter,
%% allowing the author to define a "short title" to be used in page headers.
\title{PCICF: A Pedestrian Crossing\\Identification and Classification Framework}

%% The "author" command and its associated commands are used to define
%%
%% the authors and their affiliations.
%% Of note is the shared affiliation of the first two authors, and the
%% "authornote" and "authornotemark" commands
%% used to denote shared contribution to the research.
% \author{Ben Trovato}
% \authornote{Both authors contributed equally to this research.}
% \email{trovato@corporation.com}
% \orcid{1234-5678-9012}
% \author{G.K.M. Tobin}
% \authornotemark[1]
% \email{webmaster@marysville-ohio.com}
% \affiliation{%
%   \institution{Institute for Clarity in Documentation}
%   \city{Dublin}
%   \state{Ohio}
%   \country{USA}
% }

% \author{Junyi Gu$^1$ \hspace{0.3cm} Beatriz Cabrero-Daniel$^1$ \hspace{0.3cm} Ali Nouri$^{1,2}$ \hspace{0.3cm} Lydia Armini$^1$ \hspace{0.3cm} Christian Berger$^1$}
% \affiliation{%
%   \institution{$^1$Department of Computer Science and Engineering,\\
% Chalmers University of Technology and University of Gothenburg\\$^2$Volvo Cars}
%   \city{Gothenburg} 
%   \country{Sweden}}
% \email{junyi.gu,beatriz.cabrero-daniel,christian.berger@gu.se, ali.nouri@volvocars.se}

\author{Junyi Gu}
\email{junyi.gu@gu.se}
\orcid{0000-0002-5976-6698}
\affiliation{%
 \institution{Chalmers University of Technology and University of Gothenburg}
 \city{Gothenburg}
 \state{}
 \country{Sweden}}

\author{Beatriz Cabrero-Daniel}
\email{beatriz.cabrero-daniel@gu.se}
\orcid{0000-0001-5275-8372}
\affiliation{%
 \institution{Chalmers University of Technology and University of Gothenburg}
 \city{Gothenburg}
 \state{}
 \country{Sweden}}

 \author{Ali Nouri}
\email{ali.nouri@volvocars.se}
\orcid{0000-0002-9634-6094}
\affiliation{%
 \institution{Volvo Cars,\\ Chalmers University of Technology and University of Gothenburg}
 \city{Gothenburg}
 \state{}
 \country{Sweden}}

\author{Lydia Armini}
\email{lydia.armini@gu.se}
\orcid{0009-0003-1949-8718}
\affiliation{%
 \institution{Chalmers University of Technology and University of Gothenburg}
 \city{Gothenburg}
 \state{}
 \country{Sweden}}

\author{Christian Berger}
\email{christian.berger@gu.se}
\orcid{0000-0002-4828-1150}
\affiliation{%
 \institution{Chalmers University of Technology and University of Gothenburg}
 \city{Gothenburg}
 \state{}
 \country{Sweden}}
% \author{Huifen Chan}
% \affiliation{%
%   \institution{Tsinghua University}
%   \city{Haidian Qu}
%   \state{Beijing Shi}
%   \country{China}}

% \author{Charles Palmer}
% \affiliation{%
%   \institution{Palmer Research Laboratories}
%   \city{San Antonio}
%   \state{Texas}
%   \country{USA}}
% \email{cpalmer@prl.com}

% \author{John Smith}
% \affiliation{%
%   \institution{The Th{\o}rv{\"a}ld Group}
%   \city{Hekla}
%   \country{Iceland}}
% \email{jsmith@affiliation.org}

% \author{Julius P. Kumquat}
% \affiliation{%
%   \institution{The Kumquat Consortium}
%   \city{New York}
%   \country{USA}}
% \email{jpkumquat@consortium.net}

%%
%% By default, the full list of authors will be used in the page
%% headers. Often, this list is too long, and will overlap
%% other information printed in the page headers. This command allows
%% the author to define a more concise list
%% of authors' names for this purpose.
\renewcommand{\shortauthors}{Gu et al.}

%%
%% The abstract is a short summary of the work to be presented in the
%% article.
\begin{abstract}
% Suggestion by CBe:
We have recently observed the commercial roll-out of robotaxis in various countries, including the USA and Germany. 
They are deployed within an operational design domain (ODD) on specific routes and environmental conditions, and are subject to continuous monitoring to regain control in safety-critical traffic situations. 
Since ODDs typically cover urban areas, robotaxis must reliably detect and interact with vulnerable road users (VRUs) such as pedestrians, bicyclists, and e-scooter riders. 
%Additionally, the implementation of artificial intelligence (AI) and machine learning (ML) in autonomous driving has generated a substantial amount of data. 
%However, while growing data is relatively simple, it is still challenging and expensive to inspect the quality of data and add valuable annotations.
%In addition, we also see a growing interest in end-to-end AI (E2E-AI) approaches that derive control actions for autonomous driving (AD) directly from their multimodal sensor data, with the help of complex AI/ML components, which do much more than simply perceive the vehicle's surroundings. 
To better handle such varied traffic situations, end-to-end AI, which directly computes vehicle control actions from multi-modal sensor data instead of only for perception, is on the rise.
%Therefore, in both variants of the AD software stack,
High quality data is needed to systematically train and evaluate such systems within their ODD.
In this work, we propose PCICF, a framework to systematically identify and classify VRU situations to support ODD's incident analysis. % focusing on pedestrian crossing events.
We base our work on the existing synthetic dataset SMIRK, and enhance it by extending its single-pedestrian-only design into the \emph{MoreSMIRK} dataset, a structured dictionary of multi-pedestrian crossing situations constructed systematically. 
%pedestrian crossing cases from the real-world PIE dataset, obtaining characteristic patterns that we match with our dictionary of crossing events in MoreSMIRK. 
We then use space-filling curves (SFCs) to transform multi-dimensional features of scenarios into characteristic patterns, which we match with corresponding entries in MoreSMIRK.
We evaluate PCICF with the large real-world dataset PIE, which contains more than 150 manually annotated pedestrian crossing videos. %, to evaluate the PCIC framework in both qualitative and quantitative perspectives. 
%identify and classify pedestrian crossing situations.
%We find that our framework  can filter pedestrian crossing sequences out of the sophisticated dataset and then classify the crossing patterns using an intuitive taxonomy and quantitative assessment. 
We show that PCICF can successfully identify and classify complex pedestrian crossing situations, even when groups of pedestrians merge or split during their crossing. 
By leveraging computationally efficient components like SFCs, PCICF has also potential to be used onboard of robotaxis for out-of-distribution (OOD) detection, for example.
%Our framework enables the systematic identification of relevant pedestrian crossing events in large-scale datasets to support the assessment of AI/ML systems during the development; due to the computationally efficient design of the individual algorithms like SFC, it has the potential even to be used onboard of such vehicles for out-of-distribution detection for example.
% We share an open-source replication package for PCIC containing its code as well as the complete MoreSMIRK dataset; due to the double-blind review policy, the links to code and data are only available upon request at the moment.
We share an open-source replication package for PCICF, including its algorithms, the complete MoreSMIRK dataset and dictionary, and our experiment results, available at \url{https://github.com/Claud1234/PCICF}.

\end{abstract}

%%
%% The code below is generated by the tool at http://dl.acm.org/ccs.cfm.
%% Please copy and paste the code instead of the example below.
%%
% \begin{CCSXML}
% <ccs2012>
%  <concept>
%   <concept_id>00000000.0000000.0000000</concept_id>
%   <concept_desc>Do Not Use This Code, Generate the Correct Terms for Your Paper</concept_desc>
%   <concept_significance>500</concept_significance>
%  </concept>
%  <concept>
%   <concept_id>00000000.00000000.00000000</concept_id>
%   <concept_desc>Do Not Use This Code, Generate the Correct Terms for Your Paper</concept_desc>
%   <concept_significance>300</concept_significance>
%  </concept>
%  <concept>
%   <concept_id>00000000.00000000.00000000</concept_id>
%   <concept_desc>Do Not Use This Code, Generate the Correct Terms for Your Paper</concept_desc>
%   <concept_significance>100</concept_significance>
%  </concept>
%  <concept>
%   <concept_id>00000000.00000000.00000000</concept_id>
%   <concept_desc>Do Not Use This Code, Generate the Correct Terms for Your Paper</concept_desc>
%   <concept_significance>100</concept_significance>
%  </concept>
% </ccs2012>
% \end{CCSXML}

% \ccsdesc[500]{Do Not Use This Code~Generate the Correct Terms for Your Paper}
% \ccsdesc[300]{Do Not Use This Code~Generate the Correct Terms for Your Paper}
% \ccsdesc{Do Not Use This Code~Generate the Correct Terms for Your Paper}
% \ccsdesc[100]{Do Not Use This Code~Generate the Correct Terms for Your Paper}

%%
%% Keywords. The author(s) should pick words that accurately describe
%% the work being presented. Separate the keywords with commas.
%% A "teaser" image appears between the author and affiliation
%% information and the body of the document, and typically spans the
%% page.
% \begin{teaserfigure}
%   \includegraphics[width=\textwidth]{sampleteaser}
%   \caption{Seattle Mariners at Spring Training, 2010.}
%   \Description{Enjoying the baseball game from the third-base
%   seats. Ichiro Suzuki preparing to bat.}
%   \label{fig:teaser}
% \end{teaserfigure}

\received{20 February 2025}
\received[revised]{12 March 2025}
\received[accepted]{5 June 2025}

%%
%% This command processes the author and affiliation and title
%% information and builds the first part of the formatted document.
\maketitle

%%%%%%%%%%%%%%%%%%%%%%%%%%%%%%%%%%%%%%%%%%%%%%%%%%%%%%%%%%%%%%%%%%%%%%%%%%%%%%%%%%%%%%%%%%%%%%%%%%%%%%%%%%%%%%
%%%%%%%%%%%%%%%%%%%%%%%%%%%%%%%%%%%%%%%%%%%%%%%%%%%%%%%%%%%%%%%%%%%%%%%%%%%%%%%%%%%%%%%%%%%%%%%%%%%%%%%%%%%%%%
%%%%%%%%%%%%%%%%%%%%%%%%%%%%%%%%%%%%%%%%%%%%%%%%%%%%%%%%%%%%%%%%%%%%%%%%%%%%%%%%%%%%%%%%%%%%%%%%%%%%%%%%%%%%%%
%%%%%%%%%%%%%%%%%%%%%%%%%%%%%%%%%%%%%%%%%%%%%%%%%%%%%%%%%%%%%%%%%%%%%%%%%%%%%%%%%%%%%%%%%%%%%%%%%%%%%%%%%%%%%%

\section{Introduction}
\label{sec:intro}
%The intelligence level has become an important aspect of the model vehicles. 
The development of information technology has changed people's opinions about vehicles, which are no longer simply mechatronic systems but software-intensive innovation platforms compared to their predecessors. 
%Particularly, software engineering plays a critical role in this transformation. 
Today, a commercial passenger car could contain more than 100 million lines of code, compared with only 50,000 lines decades ago~\cite{ThisSpectrum}, depicting the challenge the automotive industry faces in maintaining system quality.
Meanwhile, the implementation of AI and ML in the automotive industry has been extensively studied in recent years, which has led to two challenges for modern vehicles: 
First, the need for large computational power, and second, the rapid increase in onboard data to feed AI/ML software. 
%Hence, we always take these two issues into account when developing our framework for pedestrian crossing classification. 

\subsection{Problem Domain \& Motivation}
Traffic event identification and classification from such growing data is an important research direction in autonomous driving (AD) and traffic safety. 
Investigating the most safety-related mishaps highlights the importance of detecting and predicting pedestrian behavior, as they are among the VRUs. 
For instance, a pedestrian was severely injured in a recent mishap involving a robotaxi, which led to the revocation of its AD deployment and testing permit~\cite{NOURI2025112555}. 
Avoiding collisions with pedestrians typically requires the highest Automotive Safety Integrity Level (ASIL) due to the high severity of potential accidents~\cite{nouri2023stpa}. 
The implementation of these systems becomes even more challenging because pedestrian behavior and appearance are less predictable compared with those of other road users. 
For example, the vehicle may encounter pedestrians even on roads such as highways, where they are not permitted. 
Hence, the development, verification, and validation of such systems require special attention. For instance, the decomposition of safety requirements with a high ASIL (i.e., ASIL D) is prescribed by the standard, which then requires diversity in the implementations of two parallel software components performing the same task.
% Among traffic safety-related scenarios, the detection of vulnerable road users (VRUs) is one of the most critical and fundamental aspects. 
Moreover, recent research is exploring how AI/ML-based software can directly calculate vehicle control parameters from multi-modal sensor data. 
However, this trend is posing important challenges to be addressed: On the one hand, the development of AI/ML algorithms relies on growingly complex neural network (NN) architectures, and on the other hand, the development and systematic assessment of such AI/ML-based software has become dramatically dependent on large-scale datasets, which need to cover sufficiently diverse traffic situations to be effective. % It is well know that manually annotating or validating the datasets is expensive. 
We focus on two research directions in this work: a) the identification and detailed classification of VRUs when they interact with vehicles, and b) the qualification of datasets for AL/ML-based software to assess their potential to be used for specific safety-related traffic events.
Our motivation is to propose and evaluate an alternative to typical NN models that rely on complex architectures and vast computational resources to extract traffic events from large-scale non-annotated datasets. 
%We aim to utilize generalized and lightweight NN models for initial object detection, which we enrich by computationally efficient approaches to conduct semantic traffic event identification and classification.   

% \begin{itemize}
%     \item vulnerable road users must be safely detected $\rightarrow$ approaches that provide us with insights in existing datasets used for evaluation of such systems that potentially use AI/ML-enabled components are needed
%     \item datasets for training and evaluating AI/ML systems continuously grow $\rightarrow$ computationally efficient methods are needed
%     \item complementary to such approaches being suitable instruments for testing (software and systems), our contribution is focusing on computional efficiency and hence, has the potential of being applicable for real-time use, which in-turn enables live monitoring of compliance with operational design domains (ODD), i.e., if, for instance, a autonomously driving vehicle at SAE Level 4 is facing a situation it has not been designed for
% \end{itemize}

\subsection{Research Goal \& Research Questions}
Safely detecting and classifying pedestrian crossing situations in video data is critical for the robustness of perception stacks in autonomous vehicles (AVs). However, pedestrian crossing identification and classification are more than simple yes/no warnings, but also provide information about scene details such as the number of pedestrians, crossing directions, and pedestrian behaviors~\cite{chi23}.

We pursue the research goal in this work to conceptualize and prototype a framework that processes camera images directly to provide a detailed description of the pedestrian crossing event.
We first construct \emph{MoreSMIRK}, a systematic dictionary of synthetic pedestrian crossing events.
For identifying and classifying real world traffic situations, we use the broadly adopted YOLO network for preliminary object detection, and then leverage on SFC to compute a domain-specific representation of filtered pedestrian crossing tracks; this representation can be thought of as a `fingerprint' of a specific crossing event having the shape of a bar code (cf.~Fig.~\ref{fig_3:autosfc}(b)). 
These fingerprints enable us to compare the similarity to find corresponding ones in our dictionary MoreSMIRK to conclude the semantic classification of pedestrian crossing events. 
We structure our research along the following three research questions that build on top of each other:

%\textbf{RQ-1:} What should be considered when producing the synthetic dataset for a specific traffic event? 

%\textbf{RQ-2:} How to create a framework for traffic event detection with modularization, adaptation, and generalization considered?

%\textbf{RQ-3:} What is the performance of the proposed framework?

\textbf{RQ-1:} What are the design decisions to create a synthetic dataset for event classification?

\textbf{RQ-2:} What are the design decisions for a traffic event identification and classification framework?

\textbf{RQ-3:} What is the performance of the proposed framework?

% TODO: extracted from below: ``research objectives of this work is to conclude the description of pedestrian crossing events in terms of the number of pedestrians and crossing patterns.''

\subsection{Contributions \& Scope}
We present a framework for multi-pedestrian crossing identification and classification. 
We summarize our contributions as follows:

\begin{itemize}[leftmargin=0.6cm]
    \item We introduce a framework consisting of pedestrian extraction, dimensionality reduction, and crossing analysis, which incorporate NNs, SFCs, and a systematically constructed crossing event dictionary as essential algorithms, respectively. 
    \item We extend the public synthetic dataset SMIRK, which focuses on single-pedestrian cases only, to multi-pedestrian scenarios and systematically construct the MoreSMIRK dataset as a benchmark to retrieve crossing events from real-world datasets.
    \item We demonstrate and evaluate the capability and prospect of our framework in identifying and classifying pedestrian crossings in real-world scenarios. Moreover, we demonstrate the framework's potential to efficiently validate and annotate large-scale datasets for AI/ML-based software. 
\end{itemize}

Our system delivers two primary advantages: a) Our novel combination of deep learning (DL) and SFCs for event identification offers computational and storage efficiency while maintaining domain-specific information, which allows us to deploy the framework in practical applications and consider even large-scale AD data for other traffic events; b) the MoreSMIRK dataset has the flexibility to define even complex pedestrian crossing scenarios and hence, the benchmark dictionary, which is based on the MoreSMIRK dataset, is versatile and applicable for other traffic event detections.   
% Our system relies on neural networks for preliminary object detection, then uses the space-filling curve to reduce the complexity of the detection results, and then concludes the detailed crossing event description. 
% Another essential component of our system is a data pool that contains the human-made multi-pedestrian crossing scenarios for comparison. 
% We extend the public synthetic dataset SMIRK to multi-pedestrian scenarios and produce the betterSMIRK dataset as the benchmark to retrieve the crossing event from the real-world datasets. 

%\subsection{Structure of the Article}
The remainder of the paper is structured as follows. 
Section \ref{sec:RL} reviews the pedestrian safety-related regulations, datasets, and detection approaches.  
Section \ref{sec:pcicf} presents the design and detailed architecture of PCICF.
Specifically, we provide the construction principles of the MoreSMIRK dataset, which is one of our work's essential contributions.
Section \ref{sec:method} presents the methodology used to evaluate PCICF.   
Section \ref{sec:result} reports on the results from the evaluation experiments and provides a discussion. 
Finally, a conclusion with outlook for future work is provided in Section \ref{sec:conclusion}.

%%%%%%%%%%%%%%%%%%%%%%%%%%%%%%%%%%%%%%%%%%%%%%%%%%%%%%%%%%%%%%%%%%%%%%%%%%%%%%%%%%%%%%%%%%%%%%%%%%%%%%%%%%%%%%
%%%%%%%%%%%%%%%%%%%%%%%%%%%%%%%%%%%%%%%%%%%%%%%%%%%%%%%%%%%%%%%%%%%%%%%%%%%%%%%%%%%%%%%%%%%%%%%%%%%%%%%%%%%%%%
%%%%%%%%%%%%%%%%%%%%%%%%%%%%%%%%%%%%%%%%%%%%%%%%%%%%%%%%%%%%%%%%%%%%%%%%%%%%%%%%%%%%%%%%%%%%%%%%%%%%%%%%%%%%%%
%%%%%%%%%%%%%%%%%%%%%%%%%%%%%%%%%%%%%%%%%%%%%%%%%%%%%%%%%%%%%%%%%%%%%%%%%%%%%%%%%%%%%%%%%%%%%%%%%%%%%%%%%%%%%%

% Moved to here to have it closer to section Methodology
\begin{figure*}[h!]
  \centering
  \includegraphics[width=0.95\linewidth]{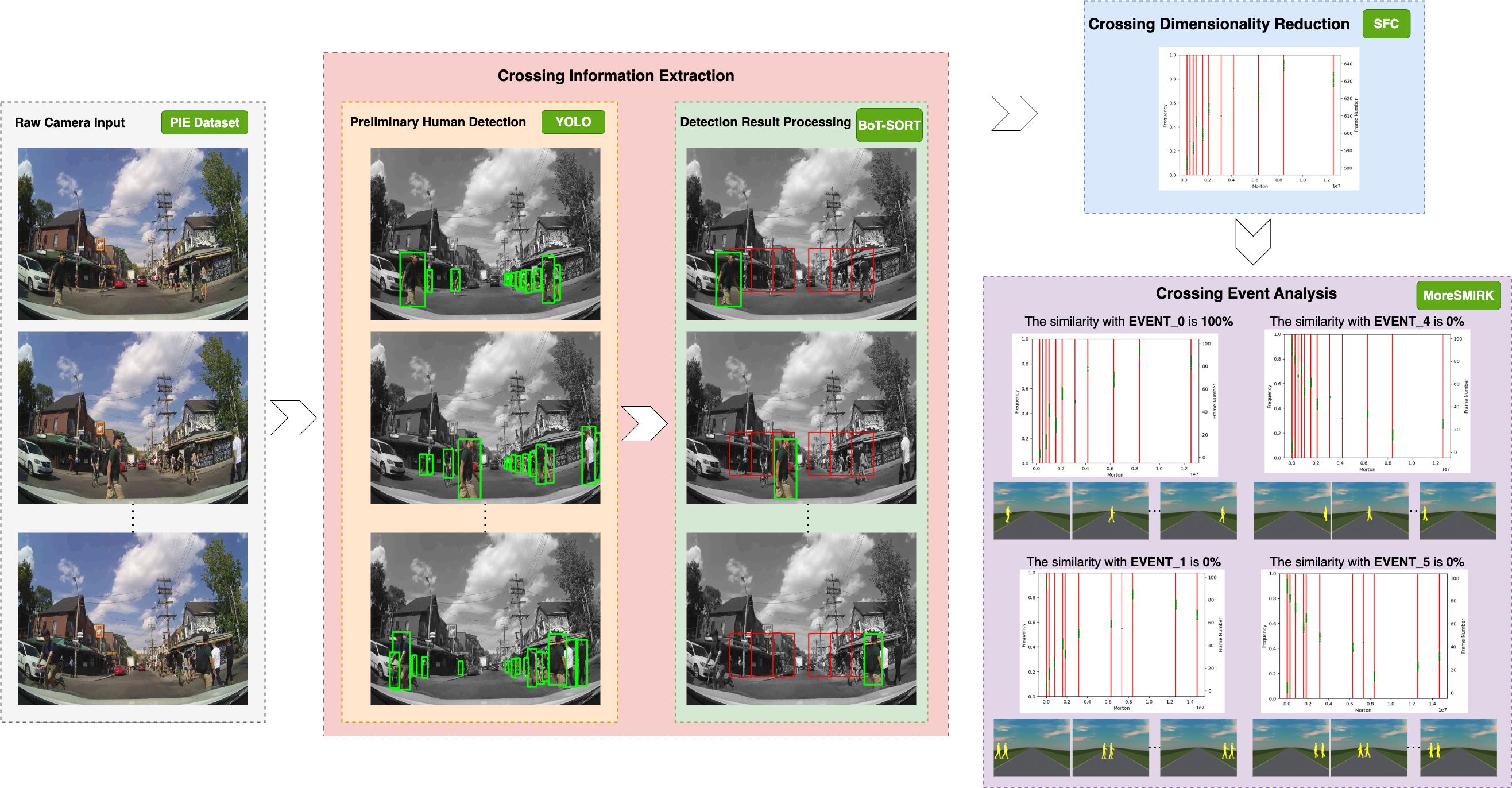}
  \caption{Overall workflow of PCICF: The raw camera input is a sequence from the PIE dataset, and the dashed rectangles represent the modules detailed in Section~\ref{sec:pcicf}. Dark-green boxes at the top indicate the datasets and algorithms used in each module. Finally, four similarity checks in crossing event analysis out of the 104 entries from MoreSMIRK are shown to obtain semantic descriptions for a crossing event. }
  \label{fig_1:diagram}
\end{figure*}

%%%%%%%%%%%%%%%%%%%%%%%%%%%%%%%%%%%%%%%%%%%%%%%%%%%%%%%%%%%%%%%%%%%%%%%%%%%%%%%%%%%%%%%%%%%%%%%%%%%%%%%%%%%%%%
%%%%%%%%%%%%%%%%%%%%%%%%%%%%%%%%%%%%%%%%%%%%%%%%%%%%%%%%%%%%%%%%%%%%%%%%%%%%%%%%%%%%%%%%%%%%%%%%%%%%%%%%%%%%%%
%%%%%%%%%%%%%%%%%%%%%%%%%%%%%%%%%%%%%%%%%%%%%%%%%%%%%%%%%%%%%%%%%%%%%%%%%%%%%%%%%%%%%%%%%%%%%%%%%%%%%%%%%%%%%%
%%%%%%%%%%%%%%%%%%%%%%%%%%%%%%%%%%%%%%%%%%%%%%%%%%%%%%%%%%%%%%%%%%%%%%%%%%%%%%%%%%%%%%%%%%%%%%%%%%%%%%%%%%%%%%

\section{Related Work} \label{sec:RL}

This section covers the state of the art of pedestrian identification and crossing classification: \autoref{sec:rw1} describes safety-related regulations, \autoref{sec:rw2} moves on to discuss existing image datasets that can be used to train AD and Advanced Driver Assistance Systems (ADAS) functions, and \autoref{sec:rw3} presents strategies to deal with large scale datasets to ensure computational efficiency in edge devices such as public road vehicles.

\subsection{Pedestrian Safety, a Requirement for AD} \label{sec:rw1}
All safety-related software systems in AVs shall avoid causing unreasonable risks to other road users, including VRUs, especially pedestrians~\cite{NOURI2025112555}. 
Hence, different types of VRUs, such as pedestrians, cyclists, or e-scooter riders, shall be considered when designing, developing, and testing perception systems in both AD and ADAS. 
Rigorous software safety assurance processes are prescribed by standards and regulations. 
For instance, ISO 21448 (Safety of the Intended Functionality), also known as SOTIF, provides requirements and recommendations for the design, verification, and validation of safety-related systems~\cite{sotif}.
SOTIF emphasizes the importance of unknown hazardous scenarios and the need to identify and mitigate them. 
This is particularly challenging for pedestrians, as specifying all combinations is impossible. 
For example, as described in the SOTIF, an unknown hazardous scenario might involve a person on a skateboard being ignored by detection algorithms because their speed exceeds the typical pedestrian walking speeds observed in the training data~\cite{sotif}.
More detailed considerations, such as diversity and sufficiency (i.e., size, age, pose) in training and testing datasets, are mentioned in ISO 8800 (Safety and Artificial Intelligence)~\cite{iso8800}.
% All safety-critical software systems, such as AVs must respect, above all other requirements, the well-being and safety of humans around them. In the automotive domain, AD and ADAS must then account for different types of VRUs such as pedestrian, cyclists, or e-scooters. 
% Software assurance techniques, especially for systems that operate in these real-time in complex real setups, are regulated by different organizations. 
% For instance, Safety Of The Intended Functionality (SOTIF) provides guidance for the design, verification and validation of safety-critical systems~\cite{sotif}, including the consideration of `reasonably foreseeable misuse by persons'. 
Complementary guidelines are provided by the European Commission's High-Level Expert Group on Artificial Intelligence: Since AVs interact with humans, a number of high-level requirements, not only during operation (i.e., transparency), but also on the data for training of the underlying AI/ML (i.e., diverse) need to be considered during development and for deployment~\cite{aiact}.

% Many AI techniques exist for pedestrian navigation classification and prediction~\cite{chi}.
% We present a framework to systematically identify pedestrian crossing events in large-scale datasets, e.g., for OOD and out of ODD detection.

\subsection{Pedestrian Image Datasets for AD training} \label{sec:rw2}
To train AI models for pedestrian identification and crossing classification, researchers often turn to already existing datasets, some of which are openly available. 
Depending on the nature of the model and the task, different types of data are used.
For instance, the Zenseact Open Dataset (ZOD) offers video frames from a high-resolution RGB front-looking camera, together with data from a LiDAR sensor~\cite{zenseact}, and covers a number of geographical areas. The large real-world dataset PIE, captured by York University in Canada, also contains more than 150 manually annotated pedestrian crossing videos.
Another well-known example is the Waymo Open Dataset, which contains recordings from five cameras in different directions side-by-side to other sensor readings and segmentation information, which is useful for scene understanding~\cite{waymo}.

While the usefulness of these datasets has been largely demonstrated, and they often include annotations such as bounding-boxes for other traffic agents, they do not reflect well the importance of focusing on vulnerable road users, such as pedestrians.
On the other hand, STCrowd is a large-scale multi-modal dataset that focuses on 3D pedestrian perception in challenging crowded scenarios~\cite{STCrowd}.

Synthetic data can also be used to train safety-critical AI models. 
Frameworks for pedestrian simulation exist~\cite{socha2022smirk}, even though not all focus on visual realism, which might hinder their use in verification processes~\cite{beaDT2}.
Some synthetic datasets focused on AD-related scenarios: For instance, SMIRK is an automatic emergency braking (AEB) system to protect pedestrians developed following the process defined in SOTIF~\cite{socha2022smirk,sotif} and has been used to conduct a systematic safety analysis of an AEB system using the AMLAS methodology~\cite{borg2023ergo}; in addition, SMIRK is also the underlying synthetic dataset focusing on single pedestrian situation used to evaluate the AEB.
Data used to train the SMIRK system is publicly available~\cite{socha2022smirk}; our work will build on it and augment it with common pedestrian motion patterns as described in Sec.~\ref{sec:pcicf}.

\subsection{Real-Time Image Analysis in Edge Devices} 
\label{sec:rw3}
As more edge cases and malfunctions in the real world are discovered, datasets for training and evaluating AVs continue to grow. 
Moreover, real-time analysis of onboard sensor data (i.e., camera feeds) is needed to enable live monitoring of surroundings of AVs and ensure compliance with pedestrian safety rules and regulations. The large sets of data as well as the need to process them in real-time, poses challenges for the software architecture and on computationally efficient processing.

Properly addressing these challenges onboard of vehicles is growingly important, where computational resources for AI systems are constrained, and connectivity might not be reliable. %~\cite{airdnd}.
We propose SFC~\cite{bader2012space} to obtain characteristic patterns in pedestrian crossings.
SFCs are mappings from high-dimensional spaces to a single-dimensional space that have practical applications in fields such as database indexing due to their ability to preserve spatial relationships between data points~\cite{bader2012space}.
Given the computational efficiency of SFCs, they have the potential even to be used onboard of vehicles, as demonstrated by Berger et al.~in 2023~\cite{Berger2023SystematicDetection,bader2012space}.

%%%%%%%%%%%%%%%%%%%%%%%%%%%%%%%%%%%%%%%%%%%%%%%%%%%%%%%%%%%%%%%%%%%%%%%%%%%%%%%%%%%%%%%%%%%%%%%%%%%%%%%%%%%%%%
%%%%%%%%%%%%%%%%%%%%%%%%%%%%%%%%%%%%%%%%%%%%%%%%%%%%%%%%%%%%%%%%%%%%%%%%%%%%%%%%%%%%%%%%%%%%%%%%%%%%%%%%%%%%%%
%%%%%%%%%%%%%%%%%%%%%%%%%%%%%%%%%%%%%%%%%%%%%%%%%%%%%%%%%%%%%%%%%%%%%%%%%%%%%%%%%%%%%%%%%%%%%%%%%%%%%%%%%%%%%%
%%%%%%%%%%%%%%%%%%%%%%%%%%%%%%%%%%%%%%%%%%%%%%%%%%%%%%%%%%%%%%%%%%%%%%%%%%%%%%%%%%%%%%%%%%%%%%%%%%%%%%%%%%%%%%

\section{PCICF: A Pedestrian Crossing Identification and Classification Framework}
\label{sec:pcicf}
%The fundamental concept of our proposed PCIC framework is an end-to-end system that first takes camera images as input, then refers to the customized pedestrian crossing datasets, and finally concludes the detailed description of the pedestrian crossing event. 
We propose PCICF, as shown in Fig.~\ref{fig_1:diagram}, as an end-to-end framework for identifying and classifying pedestrian crossing events. 
Such events can be characterized as (a) spanning over a certain amount of time (i.e., consecutive input frames in our case), and (b) the crossing event itself needs to fulfill the semantic properties as starting either on the left or right hand side of the road, and ending on the opposite side, (i.e., crossing the field of view during the selected frame sequence). 
In PCICF, camera images are fed as input to extract individual pedestrian tracks (labeled as `PIE Dataset' in Fig.~\ref{fig_1:diagram}, cf.~Section \ref{sec:method} for evaluation details), which are transformed to event-specific fingerprints that intuitively take the form of barcodes with the help of SFCs (in Fig.~\ref{fig_1:diagram}). 
We match these barcodes with those from our systematically constructed dictionary, MoreSMIRK, to obtain detailed explanations of the particular pedestrian crossing events (labeled as `MoreSMIRK' in Fig.~\ref{fig_1:diagram} ). 
In the following, we describe the relevant modules of PCICF in detail.

\subsection{Constructing the MoreSMIRK Dataset}
\label{subsec:moresmirk}
We use a look-up dictionary of known patterns for pedestrian crossing events. 
This dictionary forms the core of PCICF when analyzing a dataset of interest to obtain semantically plausible explanations for potential pedestrian crossing events. 
We based the systematic construction of MoreSMIRK on the original SMIRK dataset~\cite{socha2022smirk}, which contains 4,928 varying single pedestrian crossing configurations, along with their respective semantic segmentation and corresponding labels. 
The SMIRK dataset was created using the ESI Pro SiVIC simulator to support the systematic analysis of an automotive safety function that initiates automatic emergency braking (AEB) for a vehicle in the event of an unexpected situation where a pedestrian is crossing the road in front of a vehicle~\cite{borg2023ergo}.
The dataset contains various synthetic pedestrians, including both male and female individuals in different visual appearances (business and casual), as well as a child. 

\begin{figure}[h]
  % \centering
  \includegraphics[width=1\linewidth,trim={0 0.25cm 0 0.5cm},clip]{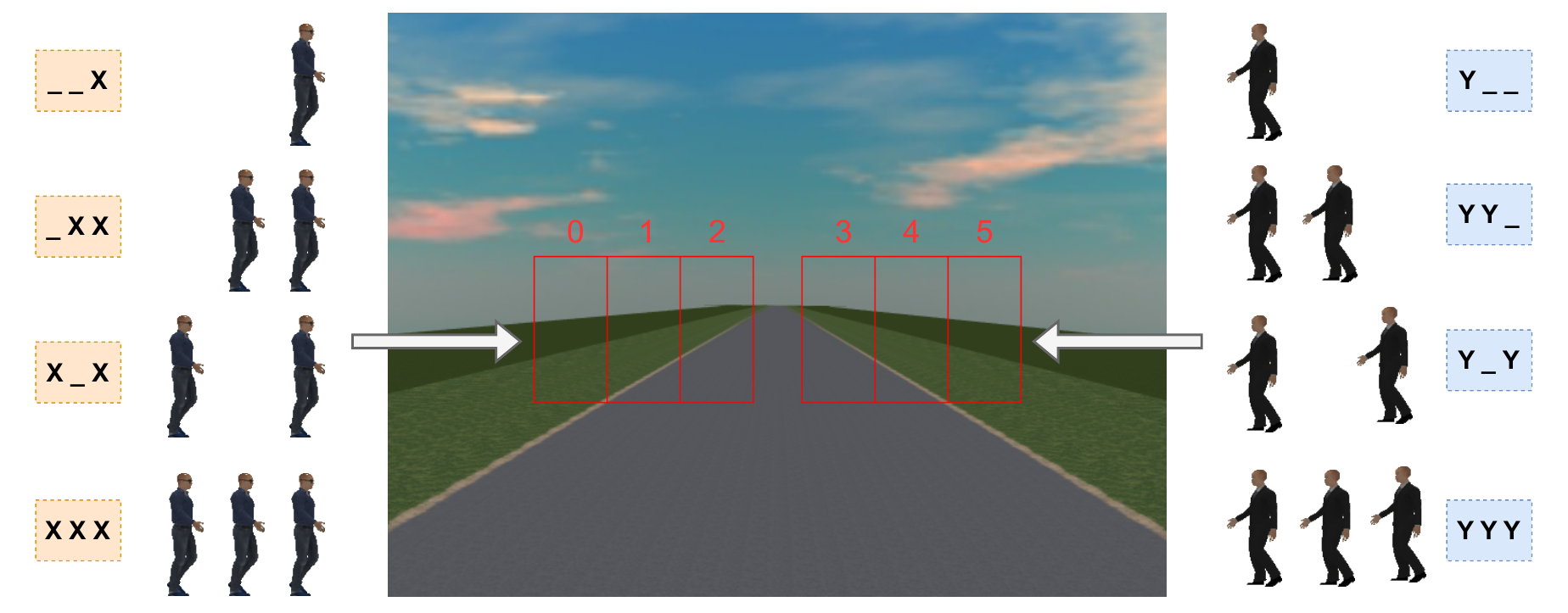}
  \caption{The illustration of configuration principles to generate the MoreSMIRK dataset. The red boxes indicate the locations of regions of interest (RoI) in the dataset.}
  \label{fig_2:moresmirk}
\end{figure}

However, the current SMIRK dataset is not sufficiently representative of what a vehicle could potentially face when approaching an inner-city intersection with traffic lights or crosswalks, as it only covers single pedestrian crossing events. 
Thus, we extend SMIRK into MoreSMIRK to improve the representativeness of pedestrian crossing situations, which are potentially present in real-world scenarios. 
Our extensions systematically add more complex pedestrian crossing configurations that were constructed according to two properties: a) initial location, and b) pedestrian grouping configuration, as shown in Fig.~\ref{fig_2:moresmirk}.
The initial location indicates where pedestrians start to cross, either from left to right or from right to left. 
We include systematically varying pedestrian configurations of up to three individuals, clearly separated and following each other when crossing a road. 
We choose this generating pattern because three people following each other with some spacing in between would almost occupy half of the street to cross, and when two groups with the same configuration would cross from both sides, the entire area in front of the ego vehicle is nearly fully covered.
All crossing events were generated as image sequences of $100$ frames, with the first pedestrian entering the scene in frame $1$ and the last pedestrian leaving in frame $100$.

As illustrated in Fig.~\ref{fig_2:moresmirk}, $X$ is the pedestrian crossing from left to right, and $Y$ represents the one crossing from right to left. 
Based on the four pedestrian group configurations, there are eight basic crossing patterns for single-directional pedestrian crossings (four from each side). 
For two-directional crossings, there are $4 * 4 = 16$ unique crossing patterns when pedestrians from two sides are moving synchronously. 
We include an additional and optional offset $\Phi$ from 0 to 5, corresponding to the number of six regions of interest (RoI) in front of the ego vehicle (cf.~red rectangle boxes in Fig.~\ref{fig_2:moresmirk}), to delay the start of the crossings for the pedestrians $X$ from the left side. 
For instance, offset $\Phi = 1$ means that when pedestrian $X$ on the left reaches the RoI grid 0, the right pedestrian $Y$ is already located at RoI grid 4, i.e., has entered the street. 
We result in $16 * 6 = 96$ crossing patterns when following this principle; in total, we provide $8 + 96 = 104$ crossing patterns in the MoreSMIRK dataset, including the pedestrian configurations without facing pedestrians starting on the other side, respectively.
Please note that mirroring the 96 two-directional sequences s is equivalent to applying the offset $\Phi$ to the pedestrian $Y$ on the right side. 
We also complement the generated scenarios with semantic ground truth annotations for all sequences in MoreSMIRK. 
Table~\ref{tab:moresmirk} shows a part of pedestrian crossing configurations. 
The complete MoreSMIRK dataset is publicly available via AI Sweden\footnote{\url{https://www.ai.se/en/ai-labs/technology-infrastructure/datasets/smirk-and-moresmirk-datasets\#moresmirk}}.

\begin{table}[]
% \centering
\caption{Part of pedestrian crossing event configurations in the MoreSMIRK Dataset. The X and Y represent the pedestrians crossing from left to right and right to left, respectively.}
\label{tab:moresmirk}
\vspace{-0.3cm}
\begin{tabular}{c|c|c|cl}
\toprule
\begin{tabular}[c]{@{}c@{}}Events\\ (E)\end{tabular} & \begin{tabular}[c]{@{}c@{}}Left Pedestrians\\ (L)\end{tabular} & \begin{tabular}[c]{@{}c@{}}Offset \\ ($
\Phi$)\end{tabular} & \begin{tabular}[c]{@{}c@{}}Right Pedestrians\\ (R)\end{tabular} &  \\ \cline{1-4}
0                                                              & \_ \_ X                                                          & N/A                                                   & \_ \_ \_                                                          &  \\
1                                                              & \_ X X                                                           & N/A                                                   & \_ \_ \_                                                          &  \\
2                                                              & X \_ X                                                           & N/A                                                   & \_ \_ \_                                                          &  \\
3                                                              & X X X                                                            & N/A                                                   & \_ \_ \_                                                          &  \\
...                                                            & ...                                                            & ...                                                   & ...                                                             &  \\
8                                                              & \_ \_ X                                                          & 0                                                     & Y \_ \_                                                           &  \\
9                                                              & \_ X X                                                           & 0                                                     & Y \_ \_                                                           &  \\
10                                                             & X \_ X                                                           & 0                                                     & Y \_ \_                                                           &  \\
11                                                             & X X X                                                            & 0                                                     & Y \_ \_                                                           &  \\
...                                                            & \textbf{...}                                                   & ...                                                   & ...                                                             &  \\
100                                                            & \_ \_ X                                                          & 5                                                     & Y Y Y                                                             &  \\
101                                                            & \_ X X                                                           & 5                                                     & Y Y Y                                                              &  \\
102                                                            & X \_ X                                                           & 5                                                     & Y Y Y                                                             &  \\
103                                                            & X X X                                                            & 5                                                     & Y Y Y                                                             & \\
\bottomrule
\end{tabular}
\end{table}

% As the original SMIRK dataset contains only single pedestrian crossing events, we used the included semantic segmentation to cut a single pedestrian's motion pattern from a given image to make the image empty from any pedestrians at all. Then, we systematically enriched this empty image by varying configurations of pedestrians by inserting the cut pedestrian's motion multiple times on the respective locations. For our three properties, we obtain 2 options for the initial locations, 2 options for the crossing directions, 4 options for group configurations (1 pedestrian as already existing in SMIRK, 2 pedestrians directly following each other, 2 pedestrians with one gap sized of the dimension of a pedestrian in between, and 3 pedestrians following each other). This generating pattern would provide us with $2 * 2 * 4 = 16$ crossing configurations; however, pedestrian groups starting on either side would move synchronously, i.e., appear as if they would mirror each other. Hence, we introduced also an extension to the first parameter (initial location), where we allow a pedestrian group configuration to start their crossing delayed of up to 5 pedestrian' widths, i.e., allowing for gaps to be added in front of that group. This increases the crossing configurations to $5 * 16 = 80$. Finally, we include the 4 pedestrian group configurations crossing solely without meeting another group of pedestrians from the other side, we ended up with $84$ pedestrian crossing configurations.

\subsection{Crossing Information Extraction}
\label{subsec:extractor}

\begin{figure*}[!]
    \centering
    \includegraphics[width=0.8\linewidth]{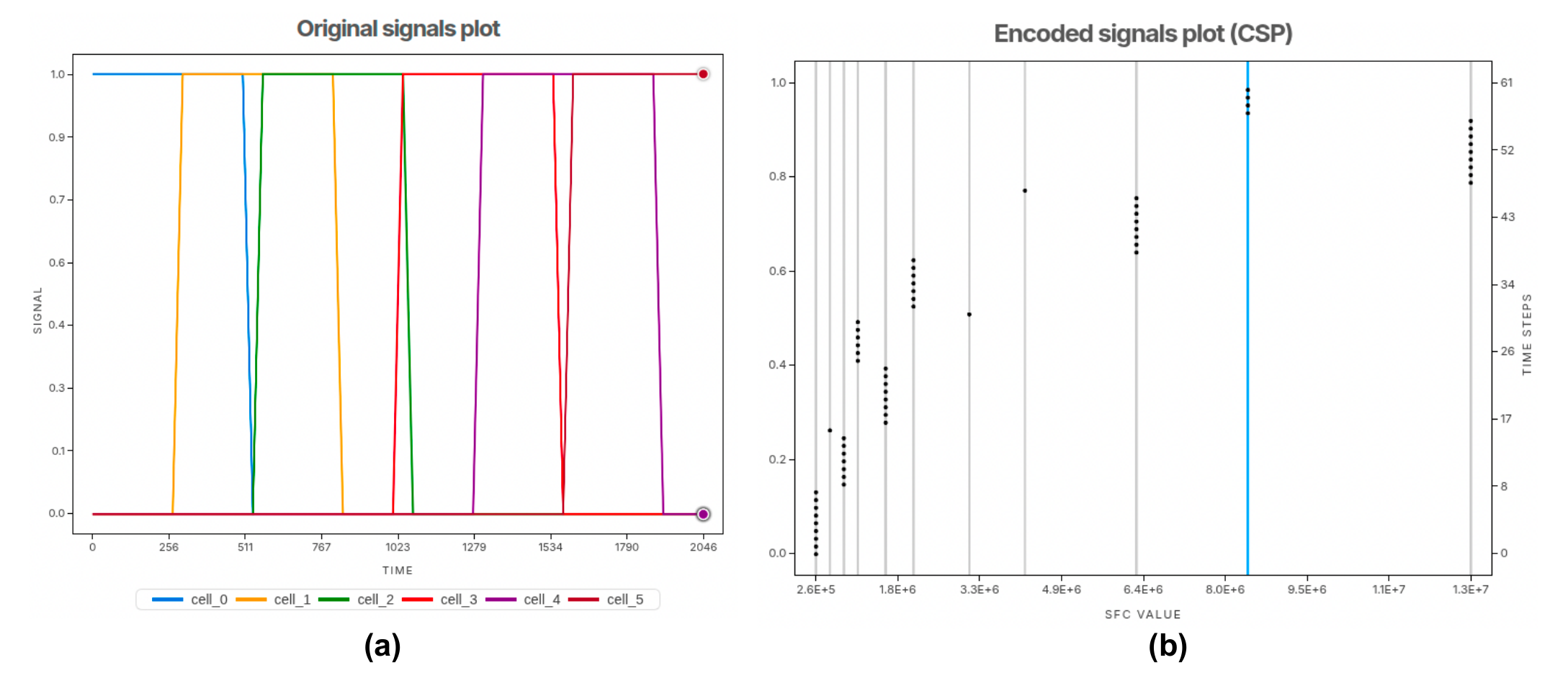}
    \caption{Data dimensionality reduction made with our AutoSFC tool$^2$ (same PIE dataset sequence as in Fig.~\ref{fig_1:diagram}): (a) shows the activation of the six RoIs (i.e., red boxes in Fig.~\ref{fig_2:moresmirk}), and (b) depicts the corresponding single-dimensional representation of the 6D-RoIs over time; the vertical stripes represent the crossing-specific fingerprint to be matched within MoreSMIRK.}
    \label{fig_3:autosfc}
\end{figure*}

PCICF consists of three individual parts: (a) preliminary pedestrian detection, (b) intermediate detection result processing, and (c) crossing event identification and classification. 
We use YOLO for preliminary pedestrian detection~\cite{yolov5} and then apply tracking and filtering to obtain relevant detections as potential candidates for pedestrian crossing sequences.
Next, we transform these sequences into their corresponding single-dimensional representations by leveraging an SFC~\cite{bader2012space} to obtain their characteristic fingerprints. 
An SFC is a curve exhibiting a repetitive pattern, which passes through every point in a multi-dimensional data space to transform it into its corresponding single-dimensional representation while, depending on the choice of the recursive pattern, preserving the original data-space's properties, such as locality between two points.
When observing such single-dimensional representations over time, we obtain characteristic stripe patterns as shown in Fig.~\ref{fig_3:autosfc} that we can match to corresponding entries within our look-up dictionary MoreSMIRK to query for matching pedestrian crossings.

%The valid pedestrian crossing sequence was summarized as an intuitive single-dimensional representation while having spatial and temporal information preserved. 
%The essence of our proposed PCICF pipeline is an end-to-end system for browsing raw datasets to retrieve critical pedestrian crossing events. 
%The classification of pedestrian crossing events focuses on the number and patterns of pedestrian crossings. 
%We leverage on the object detection neural network YOLO~\cite{yolov5} for initial human detection. % as pedestrian behavior identification and scene understanding are not in the scope of our PCIC framework.  

For the first part, preliminary pedestrian detection, we use YOLO detections in the form of bounding-boxes to filter pedestrian behavior of interest. 
YOLO is one of the most popular choices for general-purpose object detection applications because it has been optimized to balance speed and accuracy. 
However, one current shortcoming of YOLO is its relatively weak performance at separating groups of objects or densely-packed objects.

Next, for intermediate detection result processing, we extract pedestrian crossings from YOLO's detections by selecting, tracking, and filtering bounding-boxes in order to identify pedestrians' bounding-boxes within a valid crossing sequence, which spans from one side to the other side of the road. % of the vehicle's field of view (FoV) to the other. 
The valid crossing sequences are marked as a RoI grid, shown as the red boxes in Fig.~\ref{fig_2:moresmirk}, serving as a reference for bounding-box tracking and filtering. 
The definition of the RoI grid is based on domain-knowledge as the most relevant pedestrian crossing events are likely to occur in this region~\cite{bouraffa2024comparing}. 
We exclude the image's edges from the RoI grid to reduce the influence of factors such as image distortion and vehicle motion (i.e., pitching and rolling) on pedestrian detection. 
%Moreover, the constant attention area of a human driver is also excluded from the RoI grid as described in our previous work~\cite{bouraffa2024comparing}. 

We utilize the BoT-SORT~\cite{aharon2022bot,bewley2016simple} algorithm to track pedestrians, which combines a baseline model (SBS-S50)~\cite{Luo_2019_CVPR_Workshops} from the open-source object identification library~\cite{he2023fastreid} with a Kalman Filter to achieve two-tier bounding-box tracking.
Moreover, BoT-SORT proposes a camera motion compensation module to address the lack of sensor-related information, which is well-suited for our objective to process raw camera data directly. 
We choose the sparse optical flow~\cite{Bouguet1999PyramidalIO} as the backbone for feature tracking and then estimate the background motion transformation from one frame to the next. 

We filter potential pedestrian crossing events by selecting pedestrian crossing sequences that span continuously from one side of the road to the other. 
We apply the following two filtering criteria to the tracking results: (a) minimum and maximum $x$ coordinates of the pedestrian bounding-box's central point must fall in the left and right pre-defined RoI clusters (red boxes in Fig.~\ref{fig_2:moresmirk}), and (b) the horizontal displacement of a pedestrian bounding-box has to be bigger than the specific threshold (half of the whole RoI region in our experiments) that we determined heuristically.
We illustrate the process of pedestrian crossing information extraction as described above in Algorithm \ref{alg_1:extraction}. Specifically, $D$ represents the pedestrian's bounding-box information, such as the $(x, y)$ coordinates, width, and height. The BoT-SORT configuration $conf$ contains the parameters for the first- and second-tier matching thresholds, the new tracking initiation threshold, and the tracking buffer frames. 

\begin{algorithm}
 \caption{Pedestrian Crossing Information Extraction.}
 \label{alg_1:extraction}
     \begin{algorithmic}[1]
\SetNoFillComment
 \renewcommand{\algorithmicrequire}{\textbf{Input:}}
 \renewcommand{\algorithmicensure}{\textbf{Output:}}
 \REQUIRE A full input image sequence $S$; BoT-SORT configuration $conf$; bounding-box coordinates of pre-defined RoI regions $R_n$; valid-pedestrian-crossing threshold $\tau$.
 \ENSURE  The bounding-boxes of valid crossing event $E$. \\
    \tcc{The detection and tracking of bounding-boxes}
    \FOR{\textit{frame $f_n$} in $S$ } 
    \STATE $D_n \gets YOLO(f_n) $
    \STATE $uniqueID \gets BoT$-$SORT(D_n, conf)$
    \ENDFOR
    \STATE $D_{track} \gets append(argwhere(D_n, uniqueID))$ \\

    \tcc{The filtering of bounding-boxes}   
    \IF{$min(x \in D_{track})$ < $R_{n/2}$ < $max(x \in D_{track})$}
        \STATE $D_{1st-criteria} \gets D_{track}$
    \ENDIF
    \FOR{$d$ in $D_{1st-criterion}$}
    \IF{$(max(x \in d)$ - $min(x \in d))$ > $\tau$ }
        \STATE $D_{2nd-criterion} \gets d$
    \ENDIF
    \ENDFOR
    \STATE $E \gets D_{2nd-criterion}$
 \end{algorithmic}
\end{algorithm}

\subsection{Crossing Dimensionality Reduction}
\label{subsec:sfc}
%Recently, accumulating data with AI is relativity simple, and extracting valuable features and adding annotations has become more critical.
%As we leverage YOLO and BoT-SORT as part of our PCICF pipeline, our bounding-box extraction and classification module directly benefits from their high processing speed. 
%One of the research objectives of this work is to conclude the description of pedestrian crossing events in terms of the number of pedestrians and crossing directions. 
The previous processing step in PCICF provides us with semantically plausible candidates for the crossing event. 
However, the fact that pedestrians may merge or split into groups when crossing is a challenge for the classification.
We need to further process such candidates to meet our research goal of not only knowing that we face a pedestrian crossing but also obtaining further descriptive information, such as type of pedestrian crossing. 
Hence, we need to match results from the previous processing step with the systematically constructed events in MoreSMIRK.

%For this step, we reduce the data dimensionality to transform the input data to the domains that concerned by our pipeline for pedestrian crossing classification. 
%Other information, such as the speed and class categorization of pedestrians, is not in the scope of this work. 

We leverage the Z-order SFC~\cite{bader2012space} to obtain fingerprints of crossing events that we use to look up potentially matching candidates in MoreSMIRK.
We calculate fingerprints by transforming the bounding-box information across an entire sequence into a single-dimensional numerical representation that encodes the pedestrian's temporal and spatial occurrences. 
We first compute the intersection-over-union (IoU) between the valid crossing pedestrian's bounding-box and pre-defined RoI grids (the green and red rectangular boxes in Fig.~\ref{fig_1:diagram}), and then, we feed the IoU values to the Z-order SFC as input to calculate the fingerprints.
Please note that overlapping pedestrians' bounding-boxes are only considered once in the process, and we apply the floor-ceiling threshold in this step to further reduce the SFC's computational complexity. 
For the entire input sequence, the SFC algorithm produces a single-dimensional numerical representation of the activation of each box in the RoI grid, in terms of temporal and spatial perspectives. 

Fig.~\ref{fig_3:autosfc} visualizes the SFC transformation process with our tool AutoSFC\footnote{\url{https://beatrizcabdan.github.io/AutoSFC/}}: Fig.~\ref{fig_3:autosfc}(a) plots the original, multi-dimensional data from the previous pedestrian crossing extraction step in PCICF, and Fig.~\ref{fig_3:autosfc}(b) depicts the corresponding, single-dimensional representation after applying the Z-order SFC over time. 
The X-axis in Fig.~\ref{fig_3:autosfc}(a) is the time of a given sequence, and the Y-axis is the normalized IoU between the bounding-boxes and RoI grids. 
The six pre-defined RoI grids are represented by different colors, and their plots show the activation order of all boxes in the RoI grid for a given pedestrian crossing sequence.
The X and Y axes in \ref{fig_3:autosfc}(b) are the Morton code and frame ID of the given sequence after dimensionality reduction, respectively. 
The SFC transformation represents spatio-temporal information of the activation of boxes in the RoI grid in a single-dimensional representation, which is traceable to the original signals.

\subsection{Crossing Event Analysis}
\label{subsec:analyser}

The final module in PCICF is to analyze the resulting fingerprints to obtain descriptions for the crossing events. 
By using SFCs to calculate such fingerprints, the identification and classification of pedestrian crossing events over time becomes the exploration of matching fingerprints within MoreSMIRK. 
The fingerprints describe spatiotemporal events in a multi-dimensional data-space as characteristic stripe patterns (CSPs) that emerge on a single-dimensional spectrum, intuitively referred to as barcodes; we use these CSPs to identify and classify pedestrian crossing events.

% Moreover, we present an online tool, AutoSFC\footnote{Visit \url{https://beatrizcabdan.github.io/AutoSFC/}}, as shown in Fig.~\ref{fig:autosfc}, which utilizes the same SFC encoding algorithm as the PCIC framework for intuitive visualization and interaction of the data dimensionality reduction process. 
% Multi-dimensional tabulated data can be uploaded to AutoSFC to visualize multiple selected signals and explore the SFC-encoded versions side by side. 
% The tool provides control over transforming the input data, including offset and scaling, as well as over specific parameters of the SFC encoding. 
% By allowing real-time adjustments and immediate visual feedback, AutoSFC enables early insights into the characteristics and patterns present in the data points.

% In our open-source implementation, we leverage Morton codes resulting from the Z-order SFC to classify the pedestrian crossing event. 
%The essential process in this step is to compare the Morton code from a real-world input sequence with the synthetically constructed dictionary of the MoreSMIRK dataset. 
%We first feed the entire MoreSMIRK dataset, comprising 104 sequences, to the PCICF pipeline to obtain the unique Morton code arrays for the corresponding pedestrian crossing events. 
First, we need to transform the 104 pedestrian crossing sequences from MoreSMIRK via SFC into their corresponding CSPs to obtain the barcodes for the different crossing events contained in our dictionary.
Please note that this step needs to be repeated only when the content of MoreSMIRK changes to retrieve the updated CSPs, and otherwise, it is a one-time generation step.
We conduct this transformation by feeding MoreSMIRK into PCICF; the synthetic pedestrian and fixed-view background in the MoreSMIRK dataset reduces false detections in our crossing information extraction step. 
Furthermore, the steady kinetic characteristics of the MoreSMIRK dataset's pedestrians, such as speed and start/stop positions, ensure that the dimensionality reduction module can accurately determine the CSPs of each crossing event. 

When analyzing the crossing event in a sequence, the final step in PCICF is to look up potentially matching spatiotemporal CSPs within MoreSMIRK.
The similarity between the testing sequence's CSP and the one from MoreSMIRK reflects the tendency of a crossing sequence towards the one contained in our systematically constructed dictionary.
In real-traffic scenarios, a pedestrian crossing sequence often exhibits multiple characteristics simultaneously. 
For instance, a group of occluded pedestrians starts crossing with different walking speeds, and hence, they split into several clusters during or at the end of the crossing. 
For such crossing sequences, the number of pedestrians is dynamic from the ego vehicle's perspective.
A novel contribution from PCICF is that, by systematically constructing varying pedestrian crossing configurations in MoreSMIRK and using fingerprints from SFC-transformed, multi-dimensional data, we can even identify and classify such ambiguous real-world pedestrian crossing sequences, as we report in our experiments (cf.~Sec.~\ref{sec:result}).
The output from the crossing event analysis step provides similarities of the testing sequence with all crossing events defined in MoreSMIRK. 
Furthermore, these similarities also represent the temporal duration of the testing sequence that falls in each pedestrian crossing classification.

\section{Methodology for Evaluating PCICF}
\label{sec:method}

After introducing the design rationale behind PCICF, we outline its evaluation methodology based on the real-world dataset PIE~\cite{rasouli2019pie}.

\subsection{The PIE Dataset}
\label{subsec:pie}

While PCICF contains MoreSMIRK as a systematically constructed dataset with synthetically generated pedestrian crossing events, we utilize the real-world PIE dataset~\cite{rasouli2019pie} for evaluation.
PIE is an openly available, large-scale dataset focusing on pedestrians in urban traffic scenes collected by York University, Canada.
The video in PIE was captured with a wide-angle camera covering the entire pedestrian crossing area, and the dataset provides continuous, pedestrian-dense sequences as MP4 videos.
The PIE dataset claimed to have addressed issues in other popular pedestrian-related datasets, such as KITTI~\cite{geiger2012we} and JAAD~\cite{rasouli2017they}, regarding the number of samples and short discontinuous chunks.

However, as the original purpose of the PIE dataset is to support pedestrian intention analysis, we reorganized and re-labeled it to make it applicable to classify pedestrian crossing events.
We manually labeled the ground truth for pedestrian crossing events based on a similar principle that is used for constructing the MoreSMIRK dataset, i.e., start/end location of pedestrians, crossing direction, and group formation, if relevant; thereby, applying an objective labeling principle. 

We provide a total of 21,327 camera frames spanning 158 sequences, classified into seven sub-categories based on the number of pedestrians and their crossing directions. 
Moreover, for complex pedestrian crossing sequences involving multiple pedestrians and directions, we provide detailed textual descriptions to indicate critical information, such as `four pedestrians in three clusters' and `two pedestrians with gap from left to right, one pedestrian from right to left'. 
Table \ref{tab:pie} presents an excerpt of the detailed properties of the PIE dataset that we used for the evaluation of PCICF.

\begin{table}[]
\caption{The properties of part of the PIE dataset used in this work (Ped, L, and R indicate pedestrian, left, and right, respectively). }
\label{tab:pie}
\vspace{-0.3cm}
\begin{tabular}{lcc}
\toprule
\multicolumn{1}{c}{Subset}       & \begin{tabular}[c]{@{}c@{}}Number of \\ frames\end{tabular} & \begin{tabular}[c]{@{}c@{}}Number of \\ sequences\end{tabular} \\ \hline
Single Ped L to R                & 6649                                                        & 55                                                             \\ \hline
Single Ped R to L                & 6769                                                        & 56                                                             \\ \hline
Multiple follow-up Peds L to R         & 1211                                                        & 9                                                              \\ \hline
Multiple follow-up Peds R to L         & 1404                                                        & 11                                                             \\ \hline
Multiple no-follow-up Peds L to R & 1859                                                        & 10                                                             \\ \hline
Multiple no-follow-up Peds R to L & 1520                                                        & 9                                                              \\ \hline
Multiple Peds both directions     & 1915                                                        & 8     \\
\bottomrule
\end{tabular}
\end{table}

\subsection{Experimental Setup for Evaluating PCICF}

%As mentioned in Section \ref{sec:intro}, one of our research objectives is to develop an efficient approach for effectively identifying and classifying pedestrian crossing events. 
%The algorithms and tools used in different modules are widely accepted due to their lightweight nature and computational efficiency, such as YOLO, NumPy, and SFC. 
We conduct all experiments on a portable computer powered by an Intel(R) Core(TM) i5-9400 6-core CPU and an NVIDIA GTX 1660 Ti GPU. 
The parameter configurations of PCICF are crucial for achieving efficient and accurate classification. 
Please note that the framework's domain-specific parameters, such as RoI location, can be adjusted based on the application context. 
In our experiments, the input image size is $640\times480$, and each RoI box is $ 120\times 60$. 
The thresholds for the two-tier bounding-box tracking in the BoT-SORT algorithm are 0.25 and 0.1, respectively.
For the bounding-box filtering module, the valid-pedestrian-crossing threshold $\tau$, representing the displacement of the pixels to be regarded as a valid crossing, is 180. 
We share all parameters from the experimental setup in our open-source GitHub repository.

The metrics used to indicate classification results are the similarities of the Morton codes for a sequence from the PIE dataset and the corresponding sequences from the MoreSMIRK-based dictionary.
The numerical values and sequential order of Morton codes represent spatial and temporal information, which are two critical characteristics when computing the similarity percentages.
We feed all 158 sequences from PIE into PCICF and report on the similarity between the manual annotations and the results after matching the fingerprints from PIE with those in MoreSMIRK. 

\subsection{Threats to Validity}
Our work consists of two parts: (a) a systematically constructed dataset containing the dictionary of pedestrian crossing events, and (b) algorithms and prototypical implementation to process input frames, extract, and match potential pedestrian crossing sequences with the aforementioned crossing event dictionary. Potential threats may originate from these two aspects of PCICF. 

The pedestrian crossing event dictionary is produced based on the MoreSMIRK dataset for precision and minimal interference, but this may also potentially result in a limited number of pedestrian event configurations. 
For instance, the size and moving speed of synthetic pedestrians in the MoreSMIRK dataset are fixed. 
However, the distance between pedestrians and the ego vehicle varies in the real world, leading to cases where multiple pedestrians are treated as a single pedestrian (i.e., when they are far away from the vehicle or within the same RoI grid), and vice versa. 
Therefore, the classification for multiple non-following pedestrians can be biased towards single or follow-up events. 
The typical examples are sequence 161, 342, and 350 in Table \ref{tab:multi_r2l}. 
The similarity of biased events is more dominant than the ground truth (90\% against 63\%, 80\% against 54\%, and 90\% against 63\%, respectively).
Furthermore, and similar to the pedestrian intention aspect of the original PIE dataset, annotating pedestrian crossing events is a subjective topic to some extent, especially for complex pedestrian crossing sequences, which can be interpreted into different event classes. 
We follow the principle of the PIE dataset's original pedestrian intention annotation to manually label its sequences with neutral descriptions of pedestrian crossing events.

For the prototypical implementation, the choice of algorithms and related parameters may affect the accuracy and efficiency of classification. For instance, YOLO's weak performance on some pedestrian crossings in Tables \ref{tab:2_l2r} and \ref{tab:2_r2l} contributes to its failure to classify sequences in which pedestrians are far away from the ego vehicle. 
To cope with this limitation, the parameters for pedestrian tracking and filtering in PCICF would need to be calibrated and fine-tuned. 
Moreover, the tracking algorithm is the bottleneck for the classification of complex multiple-pedestrian both-directional crossings due to the heavy conclusions and background objects.  
These domain-specific nuances pose challenges for generalizability. 

%%%%%%%%%%%%%%%%%%%%%%%%%%%%%%%%%%%%%%%%%%%%%%%%%%%%%%%%%%%%%%%%%%%%%%%%%%%%%%%%%%%%%%%%%%%%%%%%%%%%%%%%%%%%%%
%%%%%%%%%%%%%%%%%%%%%%%%%%%%%%%%%%%%%%%%%%%%%%%%%%%%%%%%%%%%%%%%%%%%%%%%%%%%%%%%%%%%%%%%%%%%%%%%%%%%%%%%%%%%%%
%%%%%%%%%%%%%%%%%%%%%%%%%%%%%%%%%%%%%%%%%%%%%%%%%%%%%%%%%%%%%%%%%%%%%%%%%%%%%%%%%%%%%%%%%%%%%%%%%%%%%%%%%%%%%%
%%%%%%%%%%%%%%%%%%%%%%%%%%%%%%%%%%%%%%%%%%%%%%%%%%%%%%%%%%%%%%%%%%%%%%%%%%%%%%%%%%%%%%%%%%%%%%%%%%%%%%%%%%%%%%

\section{Results}
\label{sec:result}
This section describes the experimental results and analysis of applying PCICF to identify and classify pedestrian crossing sequences. 
We utilize PCICF to analyze a real-world dataset and then compare the similarity between of those results with the systematically constructed pedestrian crossing dictionary, MoreSMIRK.
We present here the basic information on environmental and parameter setup, along with some insights from the experimental results. 

\begin{figure}
    \centering
    \includegraphics[width=1\linewidth,trim={0 0.25cm 0 0.5cm},clip]{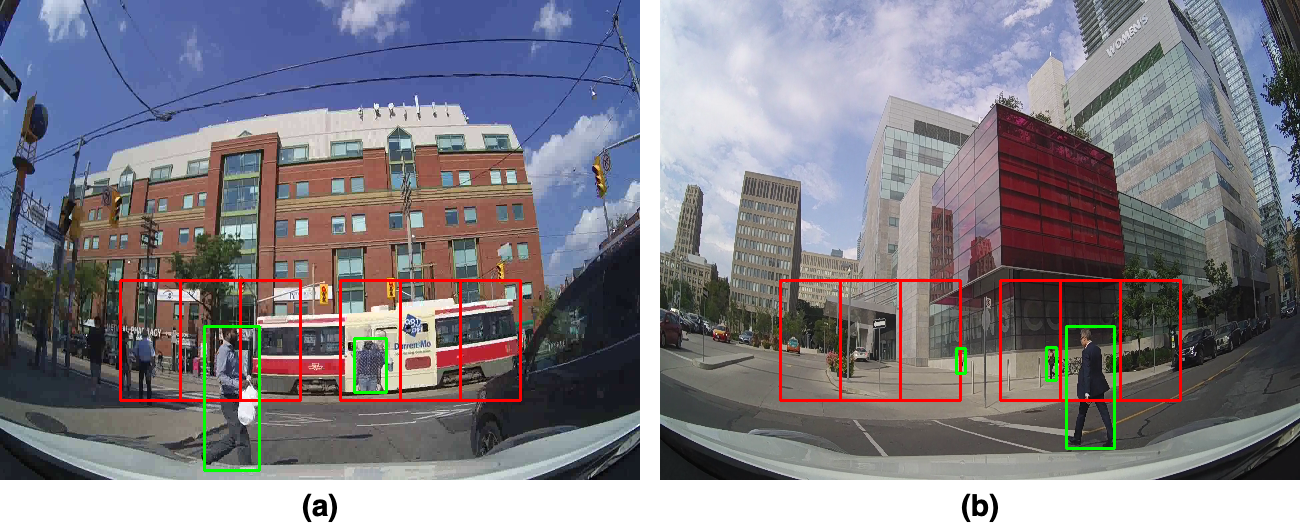}
    \vspace{-0.3cm}
    \caption{Misclassifications of single-directional crossing for single pedestrian: The ground truth for (a) and (b) are `\_ \_ X; N/A; \_ \_ \_' and `\_ \_ \_; N/A; Y \_ \_', respectively. }
    \label{fig:false-detection}
\end{figure}

As shown in Table \ref{tab:pie}, the majority of the PIE dataset consists of single pedestrian crossings, which is partially due to the real traffic at locations where the dataset was collected. 
For the scenarios of single pedestrian crossing from left to right and from right to left, corresponding to event 0 and event 4 in the MoreSMIRK dataset, respectively, our PCICF achieves accuracies of 85\% and 80\% to precisely map the PIE dataset sequences to corresponding events in MoreSMIRK. 
The primary reason for false classifications is that pedestrians at a distance are not correctly filtered or detected. 
Fig.~\ref{fig:false-detection} shows two typical false detections for single pedestrian crossing sequences. 
Fig.~\ref{fig:false-detection} (a) is the single pedestrian crossing from left to right. 
The human printed in the billboard on the tram is falsely detected; thus, PCICF returns the higher similarity for event 8 (`\_ \_ X; N/A; Y \_ \_ ' in Table \ref{tab:moresmirk}) at 66\% than the ground truth event 0 (`\_ \_ X; N/A; \_ \_ \_') at 54\%. 
Fig.~\ref{fig:false-detection} (b) is the sequence that crosses from right to left. 
The far-away pedestrian detection results in the similarity for event 5 (`\_ \_ \_; N/A; Y Y \_') at 70\% against the ground truth event 4 (`\_ \_ \_; N/A; Y \_ \_'), which is at 54\%.

Table \ref{tab:2_l2r} and Table \ref{tab:2_r2l} show PCICF's classification results for sequences of multiple follow-up pedestrians crossing from left to right, and right to left, respectively. 
We invoke a series of criteria to filter PCICF's results for multi-pedestrian single-directional crossing events: (a) only consider the similarity over 50\%; (b) discard other results if the similarity of one event is 100\%; (c) pick the first two events with the highest similarity percentages. 
Most of the follow-up crossing sequences involve only two pedestrians, PCICF returns the highest similarity with event 1 (`\_ X X; N/A; \_ \_ \_') and event 5 (`\_ \_ \_; N/A; Y Y \_'), which corresponds to their crossing event ground truth. 
However, as discussed in Section \ref{subsec:pie}, real-world pedestrian crossing sequences can exhibit characteristics of multiple events simultaneously, i.e., pedestrians split or merge during the crossing from the ego vehicle's visual view.
For these sequences with dynamic events (sequence 037 and 430 in Table \ref{tab:2_l2r}, or 008 and 009 in Table \ref{tab:2_r2l}), PCICF can indicate a relatively high similarity with another event. 
For sequences containing more than two pedestrians, PCICF classifies them into events that are more consistent with real-world situations.
In Table \ref{tab:2_r2l}, sequence 045 has four pedestrians, PCICF shows a dominating 81\% similarity of event 7 (`\_ \_ \_; N/A; Y Y Y'); the multiple pedestrians in sequence 128 are from far to close in vehicle's ego view, such change equals to the occupation of RoI grids from one to two, thus, PCICF returns a 54\% similarity of event 4 (`\_ \_ \_; N/A; Y \_ \_') and 50\% of event 5 (`\_ \_ \_; N/A; Y Y \_'). 

% The typical example is the sequence 430 in Table \ref{tab:2_l2r}; the human annotator labels the detected 347 frames sequence as `two pedestrians merge to one then split to two', and our framework returns 90\% similarity with event 0 (`\_ \_ X; N/A; \_ \_ \_') and 80\% with event 1 (`\_ X X; N/A; \_ \_ \_'). 
\begin{table}[]
\caption{Follow-up pedestrians (peds) crossing from left to right. PCICF results show the similarity of the event.  }
\label{tab:2_l2r}
\vspace{-0.3cm}
\begin{tabular}{llll}
\hline
Seq. ID & \begin{tabular}[c]{@{}l@{}}Start-End \\ Frame\end{tabular} & \begin{tabular}[c]{@{}l@{}}PCICF \\ Results \end{tabular}     & Manual Annotation                                                                          \\ \hline
003     & 598-690                                                    & \begin{tabular}[c]{@{}l@{}}E1: 70\% \\ E3: 54\%\end{tabular} & two peds                                                                                        \\ \hline
012     & 160-230                                                    & \begin{tabular}[c]{@{}l@{}}E1: 80\% \\ E3: 63\%\end{tabular} & two peds merge to one                                                              \\ \hline
021     & 139-200                                                    & \begin{tabular}[c]{@{}l@{}}E0: 54\%\\ E1: 70\%\end{tabular}  & two peds                                                                                       \\ \hline
034     & 463-602                                                    & N/A                                                         & far away, challenging                                                                      \\ \hline
036     & 232-479                                                    & \begin{tabular}[c]{@{}l@{}}E0: 63\%\\ E1: 50\%\end{tabular}  & two peds                                                                                        \\ \hline
037     & 434-637                                                    & E0: 54 \%                                                   & two peds merge to one                                                              \\ \hline
276     & 22-87                                                      & \begin{tabular}[c]{@{}l@{}}E0: 63\% \\ E1: 70\%\end{tabular} & two peds                                                                                        \\ \hline
292     & 281-400                                                    & \begin{tabular}[c]{@{}l@{}}E0: 72\% \\ E1: 80\%\end{tabular} & two peds merge to one                                                              \\ \hline
430     & 192-538                                                    & \begin{tabular}[c]{@{}l@{}}E0: 90\%\\ E1: 80\%\end{tabular}   & \begin{tabular}[c]{@{}l@{}}two peds merge to \\ one then split to two\end{tabular} \\ \hline
\end{tabular}
\end{table}
\begin{table}[]
\caption{Follow-up pedestrians (peds) crossing from right to left. PCICF results show the similarity of the event.}
\label{tab:2_r2l}
\vspace{-0.3cm}
\begin{tabular}{llll}
\hline
Seq. ID & \begin{tabular}[c]{@{}l@{}}Start-End \\ Frame\end{tabular} & \begin{tabular}[c]{@{}l@{}}PCICF \\ Results\end{tabular}       & Manual Annotation         \\ \hline
008     & 571-697                                                    & \begin{tabular}[c]{@{}l@{}}E4: 81\% \\ E5: 70\%\end{tabular} & one ped splits to two     \\ \hline
009     & 50-210                                                     & \begin{tabular}[c]{@{}l@{}}E4: 81\% \\ E5: 70\%\end{tabular}  & one ped splits to two     \\ \hline
011     & 621-731                                                    & E5: 100\%                                                      & two peds merge to one    \\ \hline
045     & 343-410                                                    & \begin{tabular}[c]{@{}l@{}}E5: 50\%\\ E7: 81\%\end{tabular}   & four peds  \\ \hline
128     & 101-232                                                    & \begin{tabular}[c]{@{}l@{}}E4: 54\%\\ E5: 50\%\end{tabular}   & four peds, far to close  \\ \hline
280     & 280-419                                                    & N/A                                                          & far away, challenging     \\ \hline
282     & 1-166                                                      & \begin{tabular}[c]{@{}l@{}}E4: 72\% \\ E5: 80\%\end{tabular}  & two peds merge to one    \\ \hline
391     & 73-175                                                     & E5: 100\%                                                      & two peds                       \\ \hline
391     & 176-271                                                    & E5: 100\%                                                      & two peds                      \\ \hline
546     & 440-590                                                    & \begin{tabular}[c]{@{}l@{}}E4: 72\%\\ E5: 70\%\end{tabular}   & two peds                      \\ \hline
\end{tabular}
\end{table}

\begin{table}[]
\caption{No-follow-up pedestrians (peds) crossing from left to right.  PCICF results show the similarity of the event.}
\label{tab:multi_l2r}
\vspace{-0.3cm}
\begin{tabular}{llll}
\hline
Seq. ID & \begin{tabular}[c]{@{}l@{}}Start-End \\ Frame\end{tabular} & \begin{tabular}[c]{@{}l@{}}PCICF \\ Results\end{tabular}       & Manual Annotation                                                                                 \\ \hline
012     & 710-790                                                    & \begin{tabular}[c]{@{}l@{}}E1: 90\% \\ E3: 63\%\end{tabular}   & \begin{tabular}[c]{@{}l@{}}three peds, two follow-up, \\ one with gap\end{tabular}                \\ \hline
048     & 1-274                                                      & \begin{tabular}[c]{@{}l@{}}E0: 54\% \\ E1: 50\%\end{tabular}  & \begin{tabular}[c]{@{}l@{}}two peds start with large gap, \\ merge to one at the end\end{tabular} \\ \hline
161     & 655-871                                                    & \begin{tabular}[c]{@{}l@{}}E3:54\%\end{tabular}               & two peds with gap                                                                                 \\ \hline
217     & 1-150                                                      & \begin{tabular}[c]{@{}l@{}}E2: 58\% \\ E3: 72\%\end{tabular}  & four peds                                                                                         \\ \hline
276     & 22-87                                                      & \begin{tabular}[c]{@{}l@{}}E0: 63\% \\ E1: 70\%\end{tabular}  & \begin{tabular}[c]{@{}l@{}}two peds start with large gap, \\ merge to one at the end\end{tabular} \\ \hline
292     & 1-280                                                      & \begin{tabular}[c]{@{}l@{}}E0: 81\% \\ E1: 50\%\end{tabular}  & three peds in different speeds                                                                    \\ \hline
313     & 1-101                                                      & \begin{tabular}[c]{@{}l@{}}E1: 50\% \\ E3: 63\%\end{tabular}     & three peds                                                                                        \\ \hline
419     & 230-584                                                    & \begin{tabular}[c]{@{}l@{}}E2: 82\% \\ E3: 81\%\end{tabular} & three peds clusters with gap                                                                      \\ \hline
445     & 501-700                                                    & \begin{tabular}[c]{@{}l@{}}E2: 76\% \\ E3: 81\%\end{tabular} & two peds clusters with gap                                                                        \\ \hline
555     & 1-144                                                      & \begin{tabular}[c]{@{}l@{}}E0: 45 \% \\ E2: 47\% \end{tabular}  & two peds with large gap                                                                           \\ \hline
\end{tabular}
\end{table}
\begin{table}[]
\caption{No-follow-up pedestrians (peds) crossing from right to left. PCICF results show the similarity of the event.}
\label{tab:multi_r2l}
\vspace{-0.3cm}
\begin{tabular}{llll}
\hline
Seq. ID & \begin{tabular}[c]{@{}l@{}}Start-End \\ Frame\end{tabular} & \begin{tabular}[c]{@{}l@{}}PCICF \\ Results\end{tabular}      & Manual Annotation                                                                           \\ \hline
024     & 49-247                                                     & \begin{tabular}[c]{@{}l@{}}E5: 70\% \\ E7: 81\%\end{tabular}  & \begin{tabular}[c]{@{}l@{}}two peds start with gap, \\ merge to one at the end\end{tabular} \\ \hline
029     & 456-599                                                    & \begin{tabular}[c]{@{}l@{}}E40: 70\% \\ E56: 80\%\end{tabular} & \begin{tabular}[c]{@{}l@{}}two peds with large gap, \\ half crossing \end{tabular}                                                                   \\ \hline
116     & 129-247                                                    & E5: 100\%                                                      & four peds in three clusters                                                                 \\ \hline
161     & 497-654                                                    & \begin{tabular}[c]{@{}l@{}}E5: 90\% \\ E7: 63\%\end{tabular}   & two peds with gap                                                                           \\ \hline
292     & 656-807                                                    & \begin{tabular}[c]{@{}l@{}}E5: 60\% \\ E6: 58\%\end{tabular}            & \begin{tabular}[c]{@{}l@{}}three peds in two clusters \\ with gap\end{tabular}                                                                     \\ \hline
342     & 564-720                                                    & \begin{tabular}[c]{@{}l@{}}E5: 80\% \\ E7:54\%\end{tabular}   & two peds with gap                                                                           \\ \hline
350     & 520-795                                                    & \begin{tabular}[c]{@{}l@{}}E5: 90\% \\ E7: 63\%\end{tabular}  & two peds with gap                                                                           \\ \hline
531     & 33-181                                                     & \begin{tabular}[c]{@{}l@{}}E7: 72\%\end{tabular} & two peds with gap                                                                           \\ \hline
543     & 522-687                                                    & \begin{tabular}[c]{@{}l@{}}E93: 70\% \\ E103: 78\%\end{tabular}  &  \begin{tabular}[c]{@{}l@{}}four peds in two clusters \\ with large gap \end{tabular}                                                         \\ \hline
\end{tabular}
\end{table}

Table \ref{tab:multi_l2r} and Table \ref{tab:multi_r2l} present the classification results for multiple pedestrians who do not follow each other to cross in a single direction. 
As mentioned in Section \ref{subsec:moresmirk}, our crossing event classification only includes up to three follow-up pedestrians from one side, as it would occupy almost half of the ego vehicle's front-facing field-of-view. 
Thus, the sequence that has more than three pedestrians, for instance, sequence 217 in Table \ref{tab:multi_l2r}, shows a dominating similarity of 72\% to event 3 (`X X X; N/A; \_ \_ \_'). 
For the manual annotation of multiple no-follow pedestrians crossing, a significant drawback is that the gap between pedestrians is represented in a subjective and relative tone. 
For instance, sequence 555 in Table \ref{tab:multi_l2r}, the human annotator describes the gap with the word `large' relative to other similar sequences. 
In contrast, PCICF reports the almost equal similarities (45\% and 47\%) of event 0 (`\_ \_ X; N/A; \_ \_ \_') and event 2 (`X \_ X; N/A; \_ \_ \_'), which indicate the gap is large enough that the sequence can be regarded as two single-pedestrian crossing sequences. 
The major false detections occur in sequences with abnormal crossing patterns, i.e., sequence 029 in Table \ref{tab:multi_r2l}, where pedestrians cross half of the road with a large gap, and sequence 543, where four pedestrians have irregularly large gaps. 
PCICF returns the classification of both-directional crossing with high offset $\Phi$. 
The reason for such false classifications is that pedestrian tracking is incorrect due to occlusion and speed changes.

\begin{table}[]
\caption{Multiple pedestrians (peds) crossing both directions. L and R indicate left and right, respectively. PCICF results show the similarity of the event.}
\label{tab:both}
\vspace{-0.3cm}
\begin{tabular}{llll}
\hline
Seq. ID & \begin{tabular}[c]{@{}l@{}}Start-End \\ Frame\end{tabular} & \multicolumn{1}{c}{\begin{tabular}[c]{@{}c@{}}PCICF \\ Results\end{tabular}}     & Manual Annotation                                                                                   \\ \hline
025     & 551-800                                                    &  E11: 31\%                & \begin{tabular}[c]{@{}l@{}}four peds L to R\\ one ped R to L\end{tabular}                      \\ \hline
031     & 1-253                                                      & E72: 80\%                 & \begin{tabular}[c]{@{}l@{}}four peds L to R\\ one ped R to L\end{tabular}                      \\ \hline
377     & 451-671                                                    & E88: 70\%                  & \begin{tabular}[c]{@{}l@{}}one ped L to R\\ two peds R to L\end{tabular}                       \\ \hline
377     & 91-357                                                     &E73: 30\%                  & \begin{tabular}[c]{@{}l@{}}two peds L to R (gap)\\ three peds R to L (gap)\end{tabular}        \\ \hline
409     & 567-822                                                    & E74: 31\%                  & \begin{tabular}[c]{@{}l@{}}two peds L to R (gap)\\ one ped R to L\end{tabular}                 \\ \hline
418     & 404-650                                                    &  E23: 35\%                & \begin{tabular}[c]{@{}l@{}}one ped L to R\\ three ped clusters R to L\end{tabular}             \\ \hline
537     & 1-240                                                      & E40: 23\%                & \begin{tabular}[c]{@{}l@{}}one ped L to R (offset)\\ two peds R to L (large gap)\end{tabular}  \\ \hline
540     & 96-276                                                     & E92: 30\%            & \begin{tabular}[c]{@{}l@{}}one pedestrian L to R\\ one pedestrian R to L (offset)\end{tabular} \\ \hline
\end{tabular}
\end{table}

Table \ref{tab:both} shows PCICF's classification results for complex multiple pedestrians crossing in irregular patterns from both directions, which is the most compelling real-world case. 
Please note that these sequences usually occur at intersections or in urban areas with busy traffic. 
We notice that factors such as background objects and occlusion pose significant challenges for PCICF's pedestrian detection and tracking modules, which result in relatively low similarities and misclassifications of offset $\Phi$. 
Instead of setting multiple criteria as the single-directional crossing scenarios, we report the event classification with the highest similarity.
For some sequences with simpler scenarios, PCICF returns a trustworthy classification. 
For instance, sequence 025 and 540's classification results are event 11 ('X X X; N/A; Y \_ \_') and event 92 ('\_ \_ X; 5; Y Y \_'), which generally correspond to the ground truth from human annotators. 

% Please note that for the two-directional multiple-pedestrian crossing sequences, especially in cases with an offset $\Phi$, their crossing-specific fingerprints always carry the corresponding spatiotemporal information from single-directional crossings (i.e., events 0 to 7 with one pedestrian) from MoreSMIRK. 
% On the contrary, the single-directional crossing sequences rarely show similarity with two-directional events (i.e., events 8 to 103 with offsets $\Phi >= 0$) as defined in MoreSMIRK.
% This characteristic attributes the first criterion for two-directional multi-pedestrian crossing classification, and then concludes the final crossing events based on the rank of the similarities.

% Please note that the real-world both-directional pedestrian crossings, especially the cases that offset  $\Phi = 1$, always carry the coincident fragment with the MoreSMIRK dataset's single-directional crossings (events 0 to 7). 
% On the contrary, the real-world single-directional pedestrian crossings rarely show similarities with the MoreSMIRK dataset's both-directional sequences (event ID higher than 7). 
% This is due to the MoreSMIRK dataset offering precise crossing events with minimal noise. 
% Based on this characteristic, we can first filter out the long pedestrian crossing sequences that involve multiple pedestrians in both-directional crossing, then rely on the highest similarity to conclude the corresponding crossing events. 
% Our framework ....

\section{Discussion}

Grounded on the aforementioned results, this section addresses the research questions. 
The first research question focused on the design decisions to create a synthetic dataset for pedestrian crossing classification. 
First and foremost, MoreSMIRK's strength lies in being a synthetic dataset that has been systematically constructed and can be systematically augmented. 
Since the focus is on pedestrian crossing scenarios, any crossing direction and group configuration is guaranteed to be covered, as opposed to previous attempts to gather video data, such as PIE~\cite{rasouli2019pie} or Waymo~\cite{waymo}, which rely on chance to capture as many pedestrian motion patterns as possible.

\begin{framed}
% \textbf{RQ-1:} What are design decisions to create a synthetic dataset for event classification?
\noindent\textbf{Answering RQ-1}: The design of a synthetic dataset for event classification should be based on a systematic exploration of the event space, and should be scalable. The goal is to allow for the augmentation with unique pedestrian crossing events, with specific crossing directions or group configurations. 
\end{framed}

It is important to note that while MoreSMIRK focuses on pedestrian motion, these principles apply to classifiers for any event type: the systematic construction of the event dictionary allows synthetic datasets to be extensible and adaptable to new configurations.
% Together with the synthetic benchmark dataset MoreSMIRK that is proposed in our work, the framework can be adopted to any traffic event detection task.

In turn, the second research question concerned the design decisions required to develop a traffic event identification and classification framework. 
To do that, the proposed framework must cover the full process from sensory data (e.g., from dashboard cameras) to an end-user-readable classification result. Given the many processing steps needed in such pipelines:

\begin{framed}
% \textbf{RQ-2:} What are the design decisions for a traffic event identification and classification framework?
\noindent\textbf{Answering RQ-2}: Traffic event identification and classification frameworks should rely on a modular and scalable processing pipeline with interchangeable and extendable components.
\end{framed}

For instance, in the proposed PCICF, the YOLO component could be replaced with another method or algorithm if a specific performance level was required for the preliminary pedestrian detection. 
Similarly, the specific SFC-based algorithm for dimensionality reduction could be replaced w another technique~\cite{bader2012space}. 

Finally, the third research question focuses on analyzing the performance of the proposed framework. 
In that regard, the goal of the proposed MoreSMIRK dataset was to mimic real-world pedestrian crossing scenarios that are challenging in terms of motion direction or group size and pattern. 
Each systematically constructed and specific motion pattern was then encoded into the MoreSMIRK dictionary and made available to PCICF. Based on the results presented in \autoref{sec:result}, we address the research question:

\begin{framed}
% \textbf{RQ-3:} What is the performance of the proposed framework?
\noindent\textbf{Answering RQ-3}: PCICF identifies single-directional crossings with high accuracy, especially single and two-follow-up pedestrians cases. However, complex multiple pedestrian two-directional crossing configurations pose challenges for PCICF.
\end{framed}

As shown in \autoref{sec:result}, the detections and classifications mostly coincide with the manual annotations of the pedestrian crossings in the case of single-directional pedestrian crossings. However, as seen in \autoref{tab:both}, more complex pedestrian crossing events have a chance for misclassification.
%This is due to the difficulty in recognizing the cell activation patterns and connecting it to the correct entry in the MoreSMIRK dictionary. 
Such events often involve heavy occlusions and background objects, however, they are also challenging to other state-of-the-art algorithms for object detection and tracking. While such algorithms use rather end-to-end AI/ML components, PCICF's focus is on computational efficiency.
%While PCICF does not offer higher recall in detecting and tracking such types of crossings, it is a computationally efficient framework and a basis for future work.
A possible extension of the PCICF pipeline could be to include a module for pedestrian crossing intention detection~\cite{chi23}. This could serve as a prior for PCICF to improve the accuracy of the classifications.

%%%%%%%%%%%%%%%%%%%%%%%%%%%%%%%%%%%%%%%%%%%%%%%%%%%%%%%%%%%%%%%%%%%%%%%%%%%%%%%%%%%%%%%%%%%%%%%%%%%%%%%%%%%%%%
%%%%%%%%%%%%%%%%%%%%%%%%%%%%%%%%%%%%%%%%%%%%%%%%%%%%%%%%%%%%%%%%%%%%%%%%%%%%%%%%%%%%%%%%%%%%%%%%%%%%%%%%%%%%%%
%%%%%%%%%%%%%%%%%%%%%%%%%%%%%%%%%%%%%%%%%%%%%%%%%%%%%%%%%%%%%%%%%%%%%%%%%%%%%%%%%%%%%%%%%%%%%%%%%%%%%%%%%%%%%%
%%%%%%%%%%%%%%%%%%%%%%%%%%%%%%%%%%%%%%%%%%%%%%%%%%%%%%%%%%%%%%%%%%%%%%%%%%%%%%%%%%%%%%%%%%%%%%%%%%%%%%%%%%%%%%

\section{Conclusions and Future Work}
\label{sec:conclusion}
In urban scenarios, ADAS and AD systems must safely interact with pedestrians crossing the road in various configurations, i.e., groups and speeds. 
Standards such as ISO 26262 and ISO 21448 guide the development and assessment of these safety-critical systems, whose robustness must be assessed systematically.

This work proposes a framework, called PCICF, to identify and classify pedestrian crossings. 
PCICF incorporates several state-of-the-art modules into its internal pipeline, including the well-known YOLO, whose detections are fed into our algorithm to semantically filter potential pedestrian crossing matches.
The pipeline also leverages SFCs to reduce pedestrian crossing sequences into their corresponding single-dimensional representations while preserving their spatiotemporal information; this allows the creation of fingerprints that visually take the shape of barcodes~\cite{Berger2023SystematicDetection}, which serve as keys in a dictionary of pedestrian crossing events called MoreSMIRK. 
The MoreSMIRK dataset, which is systematically constructed and publicly available, is used in the last module of the PCICF pipeline to obtain the final classification.

To evaluate the proposed PCICF, we rely on the large-scale, real-world dataset PIE, which contains multiple pedestrian crossing examples. 
In this dataset, PCICF achieves a maximum accuracy of 85\% in scenarios with dominating unique classifications, such as uni-directional crossings of a single pedestrian. 
On the other hand, for scenarios with ambiguous group patterns, PCICF does not provide a unique match, but instead reports the likelihoods for each crossing event type as defined in MoreSMIRK. 
Thanks to this, PCICF can also identify sub-patterns (i.e., multiple single-pedestrian crossings) within such ambiguous crossing configurations.

It is important to note that while the accuracy of PCICF, specially in challenging scenarios, might not surpass other state-of-the-art solutions, the computational efficiency of the individual algorithms in PCICF, like SFC, make it usable onboard of AVs, which typically have limited computational resource for OOD detection, for example (i.e., checking whether a configuration is not covered in the look-up dictionary). 
Moreover, given that the look-up dictionary is constructed systematically, as described in \autoref{subsec:moresmirk}, more configurations and motion patterns could be added to the look-up dictionary to expand the OOD detector.

As previously discussed, complex two-directional multiple pedestrian scenarios, common in crowded intersections, still pose challenges for PCICF. 
Thus, refining the pedestrian crossing event dictionary by systematically augmenting the range of pedestrian motion patterns, and improving the pedestrian tracking will be the focus of future work. 
For example, additional vulnerable road user types, such as e-scooter riders and people with disabilities, should be included in future editions of MoreSMIRK and PCICF to increase the inclusiveness of technology in our transportation systems.
These improvements are planned be added to the open-source PCICF on GitHub, and to MoreSMIRK, hosted at AI Sweden, to foster further important research around VRUs.

\begin{acks}
This work has been partially supported by the Swedish Foundation for Strategic Research (SSF), grant number FUS21-0004 SAICOM, Swedish Research Council (VR) under grant agreement 2023-03810, and the Wallenberg AI, Autonomous Systems and Software Program (WASP) funded by the Knut and Alice Wallenberg Foundation. Parts of the computations and data handling was enabled by resources provided by the National Academic Infrastructure for Supercomputing in Sweden (NAISS), partially funded by the Swedish Research Council through grant agreement no. 2022-06725.
\end{acks}

\subsection*{\normalsize Disclaimer}
The views and opinions expressed are those of the authors and do not necessarily reflect the official policy or position of Volvo Cars.
%%
%% The next two lines define the bibliography style to be used, and
%% the bibliography file.
\bibliographystyle{ACM-Reference-Format}
\bibliography{bib}

\end{document}